%% file: manuscript.tex
\documentclass[10pt,journal,final,twoside,twocolumn]{IEEEtran}
\ifCLASSOPTIONcompsoc
\else
\fi
\ifCLASSINFOpdf
  \usepackage[pdftex]{graphicx}
  \graphicspath{{figures/}}
\else
  \usepackage[dvips]{graphicx}
  \graphicspath{{figures/}}
\fi
\usepackage[cmex10]{amsmath}
\usepackage{array}

\ifCLASSOPTIONcompsoc
\usepackage[tight,normalsize,sf,SF]{subfigure}
\else
\usepackage[tight,footnotesize]{subfigure}
\fi
\usepackage{psfrag}
\usepackage{amssymb}

\usepackage{ctable}
\usepackage{booktabs}
\usepackage{tabularx}
\usepackage{caption}
\usepackage{algorithm}
\usepackage{algpseudocode}
\floatname{algorithm}{Code}
\usepackage{color}
\usepackage{listings}

\usepackage{balance}

\input{macro}

\begin{document}
%
\title{Regularization and Kernelization of\\the Maximin Correlation Approach}
%
%
%
%

\author{Taehoon~Lee, Taesup~Moon~\IEEEmembership{Member, IEEE}, Seung~Jean~Kim~\IEEEmembership{Member, IEEE}, and\\Sungroh~Yoon,~\IEEEmembership{Senior~Member, IEEE}%

\IEEEcompsocitemizethanks{\IEEEcompsocthanksitem T. Lee and S. Yoon are with the Department of Electrical and Computer Engineering, Seoul National University, Seoul 08826, Korea.
E-mail: sryoon@snu.ac.kr
\IEEEcompsocthanksitem T. Moon is with the Department of Information and Communication Engineering, Daegu-Gyeongbuk Institute of Science and Technology (DGIST), Daegu 42988, Korea.~
\IEEEcompsocthanksitem S. Kim was with Citi Capital Advisors, New York, NY 10013, USA.~
}
\thanks{Manuscript received March, 2016.}}

%
%

\markboth{IEEE ACCESS,~Vol.~XX, No.~X, January~20XX}%
{Lee \MakeLowercase{\textit{et al.}}: Regularization and Kernelization of A Maximin Correlation Approach}
%



\IEEEcompsoctitleabstractindextext{%
\begin{abstract}
\input{abstract}

\end{abstract}

}

\maketitle

\IEEEdisplaynotcompsoctitleabstractindextext

%
\IEEEpeerreviewmaketitle

\section{Introduction}
\input{introduction}

\section{Maximin Correlation Approach (MCA)}
\input{background}

\section{Proposed R-MCA Methdology}\label{s-methods}
\input{method}

\section{Experimental Results and Discussion}\label{s-app}
\input{experiment}

\section{Conclusion}
\input{conclusion}

\ifCLASSOPTIONcompsoc
  \section*{Acknowledgments}
\else
  \section*{Acknowledgment}
\fi
This work was supported in part by the National Research Foundation (NRF) grants funded by the Korean Government Ministry of Engineering, Science and Technology (MEST) (No. 2011-0009963) and in part by the research grants from SAP Labs and Samsung Electronics Co., Ltd.
%
%
%



\balance

\bibliographystyle{IEEEtran}
\bibliography{references}
\end{document}

%% file: macro.tex
\def\eg{{\it e.g.}}
\def\ie{{\it i.e.}}
\def\reals{{\mathbb{R}}}
\def\setX{\mathcal{X}}
\def\b1{\mathbf{1}}
\def\bu{\mathbf{u}}
\def\bv{\mathbf{v}}
\def\bw{\mathbf{w}}
\def\bx{\mathbf{x}}

\def\bX{\mathcal{X}}
\newcommand{\all}[2]{\quad #1=1,\ldots,#2}

%% file: abstract.tex
Robust classification becomes challenging when each class consists of multiple subclasses. Examples include multi-font optical character recognition and automated protein function prediction. In correlation-based nearest-neighbor classification, the maximin correlation approach (MCA) provides the worst-case optimal solution by minimizing the maximum misclassification risk through an iterative procedure. Despite the optimality, the original MCA has drawbacks that have limited its wide applicability in practice. That is, the MCA tends to be sensitive to outliers, cannot effectively handle nonlinearities in datasets, and suffers from having high computational complexity. To address these limitations, we propose an improved solution, named regularized maximin correlation approach (R-MCA). We first reformulate MCA as a quadratically constrained linear programming (QCLP) problem, incorporate regularization by introducing slack variables in the primal problem of the QCLP, and derive the corresponding Lagrangian dual. The dual formulation enables us to apply the kernel trick to R-MCA so that it can better handle nonlinearities. Our experimental results demonstrate that the regularization and kernelization make the proposed R-MCA more robust and accurate for various classification tasks than the original MCA. Furthermore, when the data size or dimensionality grows, R-MCA runs substantially faster by solving either the primal or dual (whichever has a smaller variable dimension) of the QCLP.

%% file: introduction.tex
Nearest neighbor (NN) classifiers \cite{nn2,nn3} are non-parametric methods that classify an object based on its distance to the nearest trained class. Owing largely to their simplicity and reasonable performance in practical problems, they have been widely used for various tasks such as image retrieval~\cite{Guillaumin09}, object tracking~\cite{Boltz09,Kalal10}, location-dependent information service~\cite{Zheng06}, and predicting stability of nucleic acid secondary structure~\cite{Turner09}.

\textcolor{black}{The main problems that arise with NN classifiers are that (1) it becomes computationally intensive to find the neighbors as the number of training samples increases and (2) the notion of nearest neighbors can break down in high-dimensional spaces.} Approaches have been proposed to reduce the computation~\cite{Wang07} and to adaptively determine nearest neighbors (even in high-dimensional spaces)~\cite{Noh12}. Template matching is another widely used technique that pre-computes a representative vector for each class and uses it to locate the nearest neighbor of an object~\cite{Efros03}. In multiple subclass classification problems, where each class consists of multiple subclasses, a template is constructed for each subclass, and then the \emph{aggregate} template of a class is created based on the subclass  templates~\cite{originalmca}.

In this paper, we consider constructing the aggregate template based on the idea of the \textit{maximin correlation approach} (MCA)~\cite{originalmca}. For correlation-based NN classification problems, it is known that MCA can provide an optimal aggregate template in that MCA iteratively maximizes the minimum correlation with the templates it represents, eventually minimizing the maximum misclassification risk. MCA was originally proposed for {multi-font optical character recognition}~\cite{Lebourgeois1997} and has been successfully applied to automated protein function prediction~\cite{Lee2010} and typography clustering~\cite{Lee2009}.

Despite the theoretical advantages of MCA, it has inherent limitations that have hindered wider applications in practice, such as susceptibility to noise and outliers, inability to handle nonlinearities in datasets, as well as high computational complexity. This paper proposes the \emph{regularized maximin correlation approach} (R-MCA), a significantly improved solution method that overcomes these limitations of the original MCA.



As opposed to the iterative method employed by the original MCA, we reformulate it as an instance of \textit{quadratically constrained linear programming} (QCLP)~\cite{socp}. The worst-case complexity of the iterative method grows quadratically as the number of objects increases. In contrast, the proposed QCLP formulation can be solved with linear complexity by the \textit{interior-point methods} (IPMs)~\cite{ipm} when coefficient matrices are positive semidefinite. 


Based on the QCLP formulation, we incorporate regularization and additional constraints that help R-MCA to find a robust representative vector even when (noisy) outliers exist. Our formulation has some resemblance to the regularization employed by the soft-margin support vector machine (SVM)~\cite{svm}.
%
We furthermore develop the Lagrangian dual of the regularized QCLP, which enables us to apply the kernel trick to effectively handle nonlinear structures possibly embedded in data.

This paper also presents experimental results that confirm the effectiveness of R-MCA on various public data sets. According to these results, the proposed R-MCA successfully delivers the following improvements:
%
\begin{itemize}
\item QCLP-based reformulation of MCA that enables acceleration, regularization and kernelization
\item Regularization to fight overfitting and outliers
\item Kernelization for discovering nonlinear structures
\end{itemize}

Note that R-MCA can contribute to devising a robust and scalable solution to not only nearest-neighbor classification but also a variety of other tasks based on finding group representatives. For such tasks, R-MCA can provide an alternative to conventional aggregates, such as centroids and medoids.




%% file: background.tex
To make this paper self-contained, we briefly introduce the mathematical formulation of the MCA to the reader. Additional details can be found in \cite{originalmca}.

Consider two non-zero vectors $\bu, \bx$ $\in \reals^m$. When $\bu$ and $\bx$ are column vectors, the centered correlation is defined as $\phi(\bu,\bx) = \bu^T \bx / ||\bu||_{\textcolor{black}{2}} ||\bx||_{\textcolor{black}{2}}$. MCA involves maximizing the objective function that is to find the worst-case value among the centered correlation between a non-zero vector $\bu$ and all of the vectors in a set $\bX\subseteq \mathbb{R}^m$. MCA can construct a template vector $\bu$ that maximizes the minimum correlation by the following formulation:
\begin{eqnarray}
\boxed{
\begin{array}{ll}
\mbox{maximize}   & \min_{\bx \in \bX} \phi(\bu,\bx) \\
\mbox{subject to} & \textcolor{black}{||\bu||_{2}} \neq 0.
\end{array}
}
\label{eq-mca}
\end{eqnarray}

The optimization (\ref{eq-mca}) is referred to as the \textit{MCA problem} (MCAP). The original MCA~\cite{originalmca} assumes that \textcolor{black}{all of the $\bx_i$'s are linearly independent,} $||\bx_i||_2 = 1$ for all $\bx_i \in \bX$, and $\bx_i^T \bx_j \geq 0$ for all $\bx_i,\bx_j \in \bX$ \textcolor{black}{(note that these assumptions are not required in the proposed R-MCA)}. An iterative solution to the MCAP was proposed in\textcolor{black}{~\cite{originalmca}}: the template vector $\bu$ is initialized to the centroid vector and is updated at each iteration to find the optimal vector $\bu^\star$. For fixed $m$ (the dimensionality), the worst-case complexity of this iterative algorithm is $O(n^2)$, where $n$ is the number of objects in $\bX$.

%% file: method.tex
\begin{figure}
\centering
\psfrag{a}[][][0.8]{$\bu$}
\psfrag{b}[][][0.8]{$\bu^T \bx_i = t$}
\psfrag{c}[][][0.8]{$\bu^T \bx_i < t$}
\includegraphics[width=\linewidth]{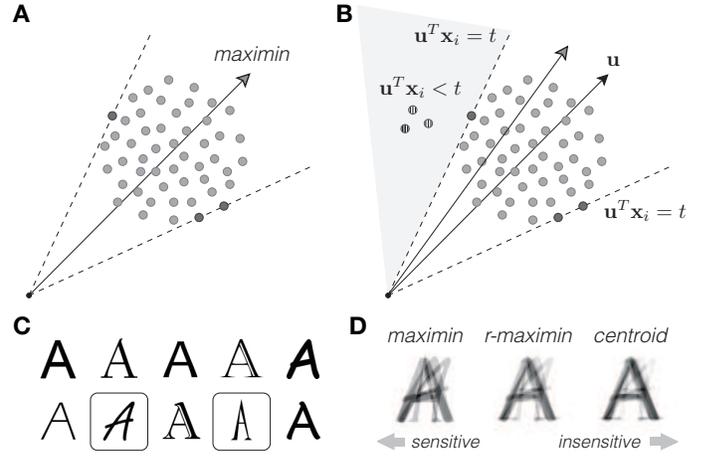}
\caption{Geometric interpretation of regularization. (a) MCA finds a vector whose direction minimizes the worst (\ie, maximum) angle between the vector and the class members. (b) Adding outliers (the shaded region) causes an abrupt swing in the traditional maximin that MCA returns. In contrast, the r-maximin that R-MCA finds is more robust to outliers. The objects on the dotted line from the origin have the minimum correlation with the template vector $\bu$. (c) The character `A' represented in 10 different fonts. (d) The three aggregate templates of (c).}
\label{f-rmca}
\end{figure}




This section presents the details of the proposed R-MCA method. 
To propose more efficient solutions to the MCAP (\ref{eq-mca}), we first formulate it as an instance of QCLP~\cite{socp}.
The QCLP formulation (\ref{eq-qclp}) enables us to find a solution using the general IPMs~\cite{ipm}, instead of the iterative method proposed in \cite{originalmca}.
The QCLP formulation also allows us to define slack variables that lead to a regularized version (\ref{eq-rmca-1}) to effectively handle outliers. From the regularized version (\ref{eq-rmca-1}), we further derive its Lagrangian dual form (\ref{eq-rmca-result}), \textcolor{black}{which reveals the structure suitable for applying the kernel trick.} 
To handle nonlinearities, we finally kernelize the dual form (\ref{eq-rmca-result}) into the kernelized R-MCA formulation (\ref{eq-kernel-rmca}).

Note that the original MCA (\ie, the version without regularization) can also be kernelized; starting from the QCLP formulation (\ref{eq-qclp}), we derive its dual form (\ref{eq-mca-result}) and the kernelized MCA (\ref{eq-kernel-mca}).





\subsection{Geometric Interpretation}

Fig.~\ref{f-rmca} shows the geometric interpretation and comparison of MCA and the proposed R-MCA, which will be formally defined in the next section. As shown in Fig.~\ref{f-rmca}(a), solving MCA is equivalent to finding a template vector whose direction minimizes the worst-case angle between the vector and class members. With no outliers, the maximin template that MCA returns represents the group reasonably.

The existence of outliers significantly degrades the performance of MCA. For instance, Fig.~\ref{f-rmca}(b) shows the scenario in which outliers are added to the data shown in Fig.~\ref{f-rmca}(a). The maximin template returned by the original MCA swings abruptly towards the outliers because MCA does not recognize outliers. In contrast, the r-maximin template returned by R-MCA takes into account the outliers, yielding a template that represents the group more reasonably.
%
%
%
%

As an example from real applications, Fig.~\ref{f-rmca}(c) shows the images of the character `A' in ten different fonts and three types of templates, each of which aims at representing the ten images as a whole. In the centroid template, the two `outlier' (boxed) fonts are averaged out and do not appear well, whereas the maximin template preserves them to some extent. For this reason, in multi-font character recognition, the maximin template, which incorporates outlier information, results in higher accuracy than the centroid template~\cite{originalmca,Lee2010}. In other applications, however, representing outliers may hurt classification accuracy. In R-MCA, we can adjust the sensitivity to outliers, providing an intermediate representation between the maximin and centroid templates (\eg, compare the three templates in Fig.~\ref{f-rmca}(c)).





\subsection{QCLP Formulation of MCA}\label{s-methods-qclp}


A simple trick allows us to reformulate (\ref{eq-mca}) as a tractable convex problem. \textcolor{black}{After normalization of input vectors, (\ref{eq-mca}) becomes} equivalent to
\begin{eqnarray}
\begin{array}{ll}
\mbox{maximize} & \min_{i=1,\ldots,n} (\bu^{T} \bx_i) \\
\mbox{subject to} & ||\bu||_{2} \leq 1.
\end{array}\nonumber
\end{eqnarray}
\textcolor{black}{The maximizer of the above maximin problem coincides with the solution of the following optimization problem:}
\begin{eqnarray}
\boxed{
\begin{array}{ll}
\mbox{maximize} & t \in \reals\\
\mbox{subject to} & \bu^T \bx_i \geq t, \all{i}{n} \\
                               & \bu^T \bu \leq 1.
\end{array}\label{eq-qclp}
}
\end{eqnarray}

The equivalent formulation (\ref{eq-qclp}) for the MCAP with a finite set $\setX$ is simple; it involves minimizing a linear function over $m+1$ variables, with $n$ linear equality constraints and one quadratic constraint. It is an instance of QCLP, a special type of optimization problem that can be solved globally and efficiently by the IPMs~\cite{ipm}.

\subsection{Regularization and Kernelization of MCA}\label{s-methods-rmca}

%
To construct a representative vector that is more robust to outliers (see Fig.~\ref{f-rmca}(b) for an example), we apply the regularization to MCA. Regularization is a popular technique to prevent overfitting. Bertsimas and Copenhaver recently described a unifying view of the connection between robustification\footnote{immunizing a statistical problem against noise in the data} and regularization~\cite{bertsimas2014characterization}.

Specifically, we introduce a non-negative `slack' variable $\xi_i$ for each object $x_i$, \textcolor{black}{which can help the optimization problem find a solution insensitive to outliers}. Using the slack variables, we can describe the regularized version of QCLP (\ref{eq-qclp}) as

\begin{eqnarray}
\boxed{
\begin{array}{ll}
\mbox{maximize}   & t - \frac{\lambda}{n} \sum\limits_{i=1}^{n} \xi_i \\
\mbox{subject to} & \bu^T \bx_i \geq t - \xi_i, \all{i}{n} \\
                               & \xi_i \geq 0, \all{i}{n} \\
                               & \bu^T \bu \leq 1
\end{array}\label{eq-rmca-1}
}
\end{eqnarray}
where $\lambda$ is a user-specified sensitivity parameter for slack variables that serves as a regularization parameter; larger $\lambda$ leads to a template vector that is more sensitive to outliers. Subsection~\ref{s-methods-effectC} presents more details of $\lambda$ and its effect on the solution of the optimization problem. Fig.~\ref{f-rmca} presents the geometric interpretation.

This formulation is similar to the optimization problem within the soft-margin support vector machine (SVM)~\cite{svm}, which is a relaxation of the original SVM. Leveraged by the regularization, the soft-margin SVM is more robust to labeling error, and we expect the proposed R-MCA to have the same advantage over the original MCA.



In order to facilitate understanding of (\ref{eq-rmca-1}), we derive its Lagrange dual~\cite{convex} problem. First of all, we define the \textit{Lagrangian L}: $ \reals \times \reals^n \times \reals^n \times \reals^n \times \reals^n \times \reals \mapsto \reals $ associated with the problem (\ref{eq-rmca-1}) as
\begin{equation}
\begin{split}
L(t, \xi, \bu, \bv, \bw, z) &= -t + \frac{\lambda}{n} \sum\limits_{i=1}^{n} \xi_i  + z ( 1 - \bu^T \bu ) \\
& - \sum\limits_{i=1}^{n} \left( v_i ( \bu^T \bx_i - t + \xi_i ) + w_i \xi_i \right)
\end{split}\label{eq-rmca-2}
\end{equation}
where $\bv = ( v_{1}, \ldots, v_{n} )^T, \bw = ( w_{1}, \ldots, w_{n} )^T \in \reals^n,$ and $z \in \reals$ are the Lagrange multipliers for the three inequality constraints of (\ref{eq-rmca-1}).
We then define the \textit{Lagrange dual function} $g$ as the minimum value of the Lagrangian over $t, \xi,$ and $\bu$:
\begin{equation}
g(\bv, \bw, z) = \inf\limits_{t, \xi, \bu} L(t, \xi, \bu, \bv, \bw, z). \label{eq-rmca-3}
\end{equation}

To calculate the infimum of the Lagrangian, we partially differentiate the Lagrangian as follows:
\begin{align}
\frac{\partial L}{\partial t} &= -1 + \sum_{i=1}^{n} v_{i} = 0 &\Rightarrow&&\sum_{i=1}^{n} v_{i} = 1 \nonumber\\
\frac{\partial L}{\partial \xi_i} &= \frac{\lambda}{n} - v_{i} - w_{i} = 0 &\Rightarrow&&\frac{\lambda}{n} = v_{i} + w_{i} \nonumber\\
\frac{\partial L}{\partial \bu} &= - \sum_{i=1}^{n} v_{i} \bx_i + 2 z \bu = 0 &\Rightarrow&&\bu = \frac{1}{2 z} \sum_{i=1}^{n} v_{i} \bx_i. \nonumber
\end{align}

From the above equalities, we can rewrite the Lagrange dual function (\ref{eq-rmca-3}) as $g(\bv, \bw, z) = - \dfrac{1}{4 z} \bv^T C \bv - z$ where $C_{ij} = \bx_i^T \bx_j $. We can consider this as a function of $z$ that is minimized when $z^\star = \sqrt{\bv^{T} C \bv} / 2 \geq 0$. Thus, we can obtain a simplified representation of $g(\bv, \bw, z)$ as $-\sqrt{ \bv^{T} C \bv }$ by substituting $z = z^\star$ into the above dual function and can finally formulate the dual problem of (\ref{eq-rmca-1}) as
\begin{eqnarray}
\begin{array}{ll}
\mbox{minimize}     & \bv^T C \bv \\
\mbox{subject to} & v_i \geq 0,\ w_i \geq 0, \all{i}{n} \\
                               & \lambda / n = v_i + w_i, \all{i}{n} \\
                               & \sum\limits_{i=1}^{n} v_i = 1.
\end{array}\label{eq-rmca-5}
\end{eqnarray}

Additionally, we can combine the top two constraints of (\ref{eq-rmca-5}) into an inequality `$\lambda / n \geq v_i \geq 0$ for all $i$', because $v_i$ and $w_i$ are complements to each other. The problem now can be described as follows:
\begin{eqnarray}
\boxed {
\begin{array}{ll}
\mbox{minimize}    & \bv^{T} C \bv \\
\mbox{subject to} & \lambda / n \succeq \bv \succeq 0 \quad \\
                                    & \mathbf{1}^T \bv = 1.
\end{array}
}\label{eq-rmca-result}
\end{eqnarray}


We can identify that (\ref{eq-rmca-result}) is a convex quadratic program (QP) since the gram matrix $C$ is positive semidefinite. Hence, when (\ref{eq-rmca-result}) has a feasible solution, by the strong duality principle~\cite{convex}, the template vector of R-MCA, $\bu^{\star}\in\reals^m$ (the primal solution), can be obtained from the solution of (\ref{eq-rmca-result}), $v^{\star}\in\reals^n$ (the dual solution), as follows:
\begin{equation}
\bu^\star = c^\star \sum\limits_{i=1}^{n} v^\star_i \bx_i\label{e:8}
\end{equation}
in which $c^{\star} = 1/{\sqrt{\bv^{\star T}C\bv^\star}}$.

\subsection{Kernelization}\label{s-methods-kernel}

The nonlinearities in input space can often be handled better in high (possibly infinite) dimensional space. The mapping to and the computation in such high-dimensional spaces can be costly, if not impossible, but when the input data are used only through inner products, we can use the so-called \emph{kernel trick} to perform implicit mapping and efficient computation. 

Inspecting the dual form of (\ref{eq-rmca-result}) immediately suggests that we can apply the kernel trick to R-MCA. Replacing the inner products in (\ref{eq-rmca-result}) with a kernel matrix $K$ yields
\begin{eqnarray}
\boxed{
\begin{array}{ll}
\mbox{minimize}     & \bv^{T} K \bv \\
\mbox{subject to} & \lambda / n \succeq \bv \succeq 0 \\
                                   & \mathbf{1}^T \bv = 1,
\end{array}
}
\label{eq-kernel-rmca}
\end{eqnarray}
where $K_{ij} = k(\bx_i,\bx_j)$ for a Mercer kernel $k$. The kernelization allows the proposed R-MCA to find template vectors from data with nonlinearities, thus extending the applicability of the R-MCA. Section~\ref{s-exp-kernel} presents more details of the kernelization and supporting experimental results.

Similarly as in kernelized R-MCA, in order to obtain a dual for MCA, we first write Lagrangian of (\ref{eq-qclp}) and its partial derivatives as follows:
\begin{align}
L(t, \bu, \bv, z) =& -t - \sum\limits_{i=1}^{n} v_i ( \bu^T \bx_i - t ) - z ( 1 - \bu^T \bu ) \nonumber\\
\frac{\partial L}{\partial t} =& -1 + \sum\limits_{i=1}^{n} v_i = 0 \nonumber\\
\frac{\partial L}{\partial \bu} =& - \sum\limits_{i=1}^{n} v_i \bx_i + 2 z \bu = 0. \nonumber
\end{align}

The Lagrange dual function of MCA thus becomes $g(\bv, z) = - \frac{1}{4 z} \bv^{T} C \bv - z$ just as in R-MCA. By inserting $z^* = \sqrt{\bv^T C \bv}/2 $ into this Lagrange dual function, we can formulate the dual of the original MCAP formulation~(\ref{eq-qclp}) as
\begin{eqnarray}
\boxed{
\begin{array}{ll}
\mbox{minimize}     & \bv^{T} C \bv \\
\mbox{subject to} & \bv \succeq 0 \\
                     & \mathbf{1}^T \bv = 1.
\end{array}
}
\label{eq-mca-result}
\end{eqnarray}

The above is the same as (\ref{eq-rmca-result}), except that the constraint $\lambda / n \succeq \bv$ is missing. In other words, the dual of R-MCA~(\ref{eq-rmca-result}) becomes the dual of MCA~(\ref{eq-mca-result}) if $\lambda >n$, hence the upper bound constraints in (\ref{eq-rmca-result}) disappears.  

The kernelized version of the original MCA can also be derived in a similar way to Section~\ref{s-methods-rmca} by replacing the dot-products in the dual quadratic program (\ref{eq-mca-result}) with a kernel:
\begin{eqnarray}
\boxed{
\begin{array}{ll}
\mbox{minimize}     & \bv^{T} K \bv \\
\mbox{subject to} & \bv \succeq 0 \\
                                   & \b1^T \bv = 1
\end{array}
}
\label{eq-kernel-mca}
\end{eqnarray}
where the constraint $\lambda / n \succeq \bv$ included in the regularized version (\ref{eq-kernel-rmca}) no longer appears.






\subsection{Analysis of $\lambda$ and a Comparison with MCA}\label{s-methods-effectC}

We elaborate on the characteristics of the template vectors obtained by R-MCA using the dual form~(\ref{eq-rmca-result}). To satisfy the constraint $\mathbf{1}^T \bv = 1$ therein, we consider the following four cases:
\begin{enumerate}
\item $[\lambda < 1]$ If the Lagrangian multipliers $v_i$'s are lower than $1/n$, the constraint $\sum v_i = 1$ cannot be satisfied. That is, (\ref{eq-rmca-result}) is not feasible if $\lambda < 1$.
\item $[\lambda = 1]$ Because $\lambda$ is the upper bound of $v_i$'s, the only solution to fit the constraint $\sum v_i = 1$ must be $v_i = 1/n$ for all $i$. In this case, $\bu^\star$ points to the same direction as the centroid of $\bx_i$'s with the scaling factor ($\bu^\star = c^\star \sum v^\star_i \bx_i = c^\star \sum \bx_i / n$).
\item $[1<\lambda <n]$ Larger $\lambda$ makes $v_i$'s less constrained; when $\lambda$ becomes large, the upper bound constraints for $v_i$'s become less restrictive for minimizing the objective $\mathbf{v}^TC\mathbf{v}$. Hence, the effect of each individual example $\mathbf{x}_i$, including the outlier, to the primal solution (\ref{e:8}) can increase as $\lambda$ increases.
\item $[\lambda \geq n]$ If $\lambda \geq n$, the upper bound constraints for $v_i$'s disappear; this follows from the fact that $\bv$ is forced to be a probability vector by the other constraints, and thus it will always satisfy the upper bound constraints when $\lambda \geq n$. By comparing (\ref{eq-rmca-result}) with the dual of the original MCA formulation (\ref{eq-qclp}) as below, we deduce that the solution of R-MCA for this case coincides with that of MCA.
\end{enumerate}

\subsection{Complexity Analysis}

Recall that $n$ corresponds to the number of objects and $m$ corresponds to the dimensionality. Since the number of iterations that is necessary for IPM to find a solution is practically constant (typically from 10 to 50)~\cite{socp}, we can see that the QCLP (\ref{eq-qclp}) can be solved in $O(nm^2 + m^3)$ flops. 
For comparison, the number of flops required for the iterative method \cite{originalmca} is either $4mnp - mp^2$ or $4n^2p - 2np^2 + mn^2$, depending on the implementation, where $p$ is the number of iterations. The empirical study in~\cite{originalmca} shows that $p$ grows nearly linearly in $n$. In result, MCA has order of $O(n^2 m)$ or $O(n^3 + mn^2)$ time complexity due to the $p=O(n)$, while the proposed QCLP formulation takes $\min ( O(nm^2 + m^3), O(n^3) )$. The computational efficiency will be demonstrated in Section~\ref{s-exp-time}.


%% file: experiment.tex

\begin{table}
\renewcommand{\arraystretch}{1.4}
\centering
\caption{Data Used in Our Experiments}\label{t-datainfo}
\begin{tabular}{lrrll}
\toprule
Name & $n$ & $m$ & Number of classes (description)\\
\midrule
KSC \cite{Kumar2002} & 5211 & 176 & 13 (land cover types)\\
MNIST \cite{Lecun1998} & 10000 & 784 & 10 (digits `0'--`9') \\
SONAR \cite{Gorman1988} & 208 & 60 & 2 (rock or mine) \\
GEO \cite{Noble08} & 606 & 30954 & 2 (ulcerative colitis patient or not) \\
3D-NUT & 272 & 3 & 2 (core or shell) \\
\bottomrule
\end{tabular}
\end{table}


We tested the proposed R-MCA methodology using the datasets listed in Table~\ref{t-datainfo}. More details about each dataset will be provided in the following subsections.

For our experiments, we implemented the proposed QCLP-based MCA and R-MCA solvers using SeDuMi software, a MATLAB toolbox for optimization over symmetric cones~\cite{Sturm98usingsedumi}. For comparison, we also prepared implementations of the original iterative solution to MCA as described in~\cite{originalmca}, the support vector machine (SVM), and the logistic regression.

\begin{figure}
\centering
\psfrag{a}[][][0.7]{centroid}
\psfrag{c}[][][0.7]{$\mathbf{\lambda=1.1}$}
\psfrag{e}[][][0.7]{$\mathbf{\lambda=1.4}$}
\psfrag{m}[][][0.7]{$\mathbf{\lambda=1.7}$}
\psfrag{n}[][][0.7]{$\mathbf{\lambda=2.0}$}
\includegraphics[width=\linewidth]{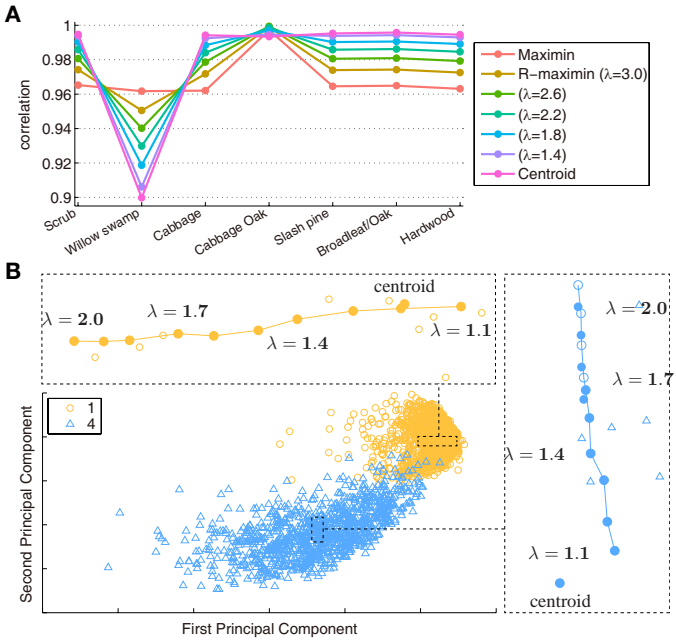}
\caption{Effects of regularization and its parameter $\lambda$. (a) The minimum correlation between aggregate templates and subclass templates of the KSC data. (b) The part of the MNIST data representing the digits `1' (yellow) and `4' (blue).}
\label{f-result1}
\end{figure}

\subsection{Effect of Regularization on Subtype Correlation}\label{ss:regularization}


To see the effect of regularization, we ran R-MCA with different values of parameter $\lambda$ on a multiple-subclass dataset and measured the variation of correlation between subclass objects and the aggregate template. We used the Kennedy Space Center (KSC) dataset~\cite{Kumar2002}, which contains 5211 vectors with 176 dimensions. Each vector represents the signal intensities of different wavelengths measured above 13 types of land covers (105--927 vectors per class). 
Based on the characterization of vegetation, these classes can be grouped into three types or `superclasses'  (upland with seven land-cover subclasses, wetland with five, and water type with one).

\figurename~\ref{f-result1}(a) shows the correlation of seven subclasses of the `upland' class with the regularized maximin aggregate templates (r-maximin) of five different $\lambda$ values (1.4, 1.8, 2.2, 2.6, 3.0). As mentioned earlier, we define $\lambda$ to manipulate the degree of regularization and can increase the best-case correlation value with the class members instead of sacrificing the worst-case correlation. To verify this effect, the curves for the non-regularized maximin and the centroid template are also presented. As expected, the curves for the r-maximin are placed between the centroid and the non-regularized maximin. 



%

\subsection{Effect of Regularization Parameter $\lambda$}\label{s-exp-effectC}

In Section~\ref{ss:regularization}, we discussed that the regularization parameter $\lambda$ works as a control knob that places the result from using the r-maximin somewhere between those from the centroid template and the non-regularized maximin template. To visualize the effects of varying $\lambda$, we utilized the MNIST database of handwritten digits~\cite{Lecun1998}. From this database, we sampled 1135 and 982 images representing the digits `1' and `4', respectively. Each sample is a $28\times28$ image that can be represented by a 784-dimensional vector. We carried out the PCA of these samples and took the first two principal components only, transforming each of them into a 2-dimensional point, as shown in \figurename~\ref{f-result1}(b). In the figure, each of the two inlets magnifies the centroid and the r-maximin along with the corresponding images for visual inspection.

Recall from Section~\ref{s-methods-effectC} that R-MCA eventually produces a centroid when $\lambda=1$. As depicted in Fig.~\ref{f-result1}(b), we tested 10 different $\lambda$ values of the interval $[1.1, 2.0]$ to draw the trajectories of the r-maximin. The centroid in the class `1' is located in the upper-right region, because most samples in `1' class are distributed in that region. However, it is necessary to shift the aggregate template toward the outliers in order to minimize worst-case classification risk. We confirmed that reducing $\lambda$ puts the regularized maximin template near the centroid, and increasing $\lambda$ yields the r-maximin close to the outliers.

\subsection{Effect of Regularization on Classification}

\begin{figure}
\centering
\includegraphics[width=\linewidth]{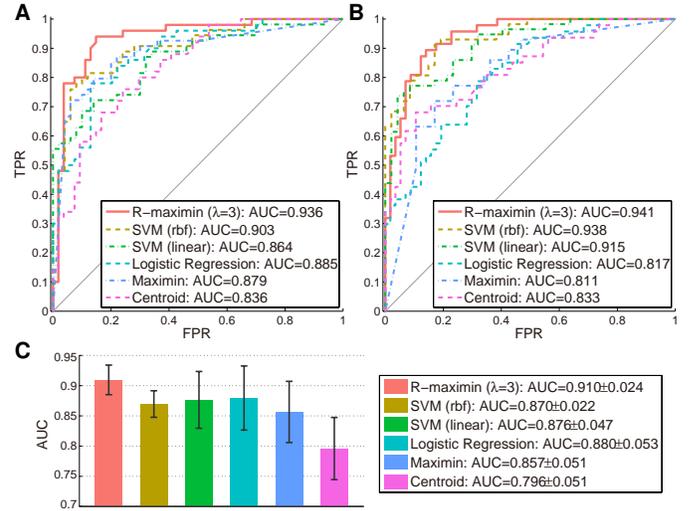}
\caption{Regularization improves classification performance using SONAR. Hyper(a, b) ROC curves from 2-fold cross validation. (c) AUC values ($\mu \pm \sigma$) from 5-fold cross vaidation.}
\label{f-result2}
\end{figure}

To see the regularization effects in the context of classification, we carried out binary classification of the SONAR data~\cite{Gorman1988}, which consist of 111 mine-reflected and 97 rock-reflected sonar signals of 60 dimensions each. For NN classification using templates, we implemented the nearest template classifier that assigns an unknown vector to the class of its nearest (r-maximin, maximin, or centroid) template. For comparison, we also tested logistic regression, the linear SVM, and the RBF kernel SVM.

According to the experimental results from using neural networks in~\cite{Gorman1988}, nonlinearities exist in the distribution characteristics of the SONAR data. We thus preprocessed the data using the kernel PCA~\cite{kernelpca} with the Gaussian kernel ($\sigma = 1$). We then tested the five different classifiers with 2 and 5-fold cross-validation. The value of $\lambda$ was determined by performing the CV with 4 different $\lambda$ values $(1.5, 2, 2.5, 3)$. The soft margin coefficient and the sigma of RBF kernel in SVM are 1 and 3, respectively.

\figurename~\ref{f-result2}(a) and (b) shows the receiver operating characteristic (ROC) curves from the first and second rounds of CV with $\lambda=3$. The average area under the curve (AUC) values are 0.94, 0.90, 0.86, 0.89, 0.88, and 0.84 for NN with the r-maximin templates, the RBF SVM, the linear SVM, logistic regression, NN with the maximin templates, and NN with the centroid templates, respectively. \figurename~\ref{f-result2}(c) also presents an average and a standard deviation of 5 runs. The r-maximin classifier achieved 3.3\% higher AUC on average than the alternatives.

With respect to these AUC values, the r-maximin classification produced the best result, whereas the performance of the original maximin classification was lower than that of the SVM. This result suggests that the regularization can indeed improve the classification accuracy for real applications with noise.
%



\subsection{Effect of Kernelization}\label{s-exp-kernel}

\begin{figure}
\centering
\includegraphics[width=\linewidth]{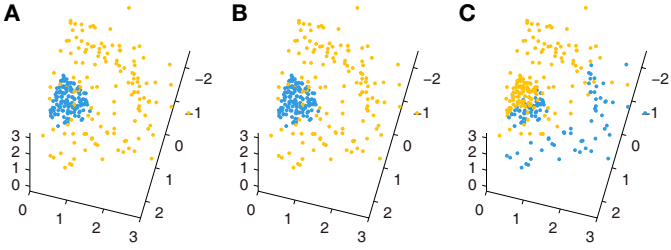}
\caption{Effect of kernelization (data: 3D-NUT). (a) The true membership (best viewed in color). (b) The membership retrieved by the proposed kernelized R-MCA. (c) The membership assigned by the R-MCA.}
\label{f-result3}
\end{figure}

Through kernelization, we expect R-MCA to become applicable to classification problems that contain complex shapes in the input space. \figurename~\ref{f-result3} shows the result from a proof-of-concept experiment using a synthetic dataset termed 3D-NUT, which was generated as follows: we sampled a point $\mathbf{x}=[x_1,x_2,x_3]$ from a trivariate normal distribution $\mathcal{N}(\mathbf{\mu}, \mathbf{\Sigma})$, where $\mathbf{\mu} = [0, 0, 0]$ and $\mathbf{\Sigma} = \mathbf{I}$. For the sake of visualization, $\mathbf{x}$ was discarded if $x_2<0$ and $x_3<0$. Otherwise, we set the membership of $\mathbf{x}$ to the `core' class if $||\mathbf{x}|| < 1$ and to the `shell' class if $||\mathbf{x}|| > 2$.

\figurename~\ref{f-result3}(a) depicts the distribution of 272 points color-coded with binary membership (either `core' or `shell' class) in the input space.
Applying the original R-MCA resulted in incorrect classification, as shown in \figurename~\ref{f-result3}(c).
In contrast, the kernelized R-MCA (radial basis kernel with $\gamma=1$) correctly separates the data points according to their membership, as shown in \figurename~\ref{f-result3}(b).

This experiment confirms that the kernelization works for R-MCA, and that we will be able to apply the kernelized version to other problems existing kernel-based methods (\eg, kernel PCA) can be applied to.


\begin{figure}
\centering
\psfrag{b}[l][][0.55]{\cite{originalmca}}
\psfrag{d}[l][][0.6]{(\ref{eq-rmca-result})}
\psfrag{f}[l][][0.6]{(\ref{eq-rmca-1})}
\psfrag{h}[l][][0.6]{(\ref{eq-mca-result})}
\psfrag{i}[l][][0.6]{(\ref{eq-qclp})}
\includegraphics[width=\linewidth]{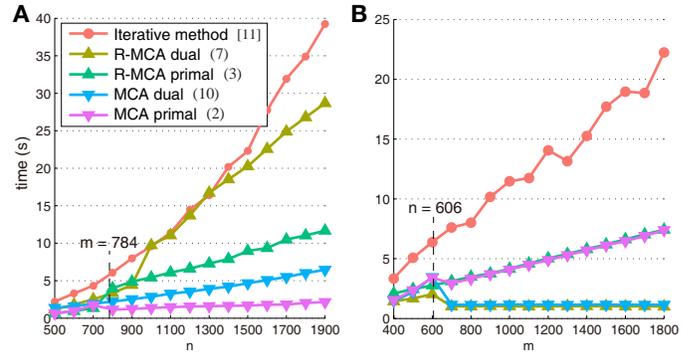}
\caption{Comparison of execution times. Each time point represents the average of ten independent runs. (a) Varying $n$ with fixed $m=784$ (data: MNIST). (b) Varying $m$ with fixed $n=606$ (data: GEO).}
\label{f-result-time}
\end{figure}

\subsection{Comparison of Execution Time with MCA}\label{s-exp-time}
We compare the runtime of the proposed QCLP-based solution and the original iterative solution~\cite{originalmca} to the maximin correlation approach. To this end, we carried out two types of experiments. One is varying the number of objects $n$ with the dimensionality $m$ fixed, and the other involves varying $m$ with $n$ fixed. We measured the runtime using a Windows 7 PC equipped with an Intel i5-3570K CPU (3.4GHz, 6MB, 5GT/s) and 16GB RAM.



\figurename~\ref{f-result-time}(a) shows the varying-$n$ fixed-$m$ case for recognizing the digit `0' in the MNIST data (fixed $m=784$). The time demand of the iterative solution remained the highest and also grew up faster than the others. As described in Section~\ref{s-methods}, there are additional inequality constraints and variables in the regularized forms [(\ref{eq-rmca-1}) and (\ref{eq-rmca-result})] in comparison with the original MCA [(\ref{eq-qclp}) and (\ref{eq-mca-result})]. Consequently, the two regularized versions require longer execution times than the unregularized ones when $n>m$, as shown in \figurename~\ref{f-result-time}(a).

The varying-$m$ fixed-$n$ case is presented in \figurename~\ref{f-result-time}(b). We used the NCBI GEO microarray dataset~\cite{Barrett2005} (the accession number: GSE11223), which provides the regional variation of gene expression in ulcerative colitis patients~\cite{Noble08}. The dataset has $m=30954$ features and $n=606$ samples (404 samples were generated by adding white Gaussian noise to the original 202 samples). Even though $m$ increases, the runtime of the dual forms [(\ref{eq-rmca-result}) and (\ref{eq-mca-result})] does not increase noticeably, because $n\times n$ quadratic programming is involved in solving the dual forms. In contrast, the time demand of solving the primal forms [(\ref{eq-qclp}) and (\ref{eq-rmca-1})] increases as $m$ grows. Consequently, if $m > n$, the $n\times n$ quadratic programming would take less time, and solving the dual forms would be better.



Note that we can observe abrupt changes in runtime from both \figurename~\ref{f-result-time}(a) and (b) at the point where $m=n$.  This originates from the design of the SeDuMi toolbox. It uses an approximation based on the Farkas' lemma~\cite{convex} and finds the solution $y \in \reals^m$ such that $A^T y = 0$ if the solution $x \in \reals^n$ does not exist for $A x \geq 0$.

In summary, the primal and dual forms should yield the same solution, and we can always solve either the original MCA or the proposed R-MCA problems faster by using the proposed QCLP formulation than using the original iterative method.
When $n>m$, using the primal forms [(\ref{eq-qclp}) and (\ref{eq-rmca-1})] will be advantageous; otherwise using the dual forms [(\ref{eq-rmca-result}) and (\ref{eq-mca-result})] will be desirable. As the primal forms and the dual forms have $O(m)$ and $O(n)$ variables, respectively, the same observations can be made from the computational complexity of SeDuMi, which is $O(x^2 y^{2.5} + y^{3.5})$~\cite{Sturm98usingsedumi} ($x$ is the number of variables, and $y$ is the number of independent inequalities). 

%% file: conclusion.tex
\label{s-conclusion}

The maximin correlation approach (MCA) was originally proposed in the context of multiple-subclass classification problems that range from the optical character recognition problem to the automated protein family prediction. The aggregate templates found by MCA work well for such applications since they can minimize the maximum misclassification risk in the correlation-based nearest-neighbor classification setup. Nonetheless, practical limitations such as susceptibility to noise, inability to handle nonlinearities consideration, and high time demand have hindered a wider application of MCA to real applications.

To address these drawbacks, we first described how to formulate the MCA as an instance of the QCLP and presented an efficient and general solution that can replace the original iterative solution. Based on this QCLP-based formulation, we further explained how to regularize and kernelize MCA in order to render it more robust to outliers and applicable to data with nonlinearities.


According to our experimental results, the proposed R-MCA successfully overcomes the limitations of the original MCA. Leveraged by the regularization, the proposed method outperformed the original MCA and the other alternatives tested in terms of classification performance. Given that the degree of regularization in R-MCA can be adjusted conveniently via a single parameter, the proposed R-MCA provides a flexible solution. In addition, we confirmed the computational benefit of the QCLP formulation and the effectiveness of kernelization in the (regularized) maximin correlation approach.

We anticipate that the kernelization and regularization of MCA will make MCA more appealing to a wider range of applications that we otherwise cannot satisfactorily analyze with the original MCA.


